\documentclass{article}

\PassOptionsToPackage{numbers, compress}{natbib}

\usepackage[final]{neurips_2019}

\usepackage[utf8]{inputenc} 
\usepackage[T1]{fontenc}    
\usepackage{hyperref}       
\usepackage{url}            
\usepackage{booktabs}       
\usepackage{amsfonts}       
\usepackage{nicefrac}       
\usepackage{microtype}      
\usepackage{tikz}
\usetikzlibrary{calc}
\usepackage{amsmath}
\usepackage{svg}

\title{Gradient Boosted Decision Tree Neural Network}

\author{%
    Mohammad Saberian \quad Pablo Delgado \quad Yves Raimond \\
    Netflix\\
    \texttt{\{esaberian, pdelgado, yraimond\}@netflix.com} \\
}
\begin{document}

\maketitle
\vspace{-.1in}
\begin{abstract}
In this paper we propose a method to build a neural network that is similar to an ensemble of decision trees. We first illustrate how to convert a learned ensemble of decision trees to a single neural network with one hidden layer and an input transformation. We then relax some properties of this network such as thresholds and activation functions to train an approximately equivalent decision tree ensemble. The final model, Hammock, is surprisingly simple: a fully connected two layers neural network where the input is quantized and one-hot encoded. Experiments on large and small datasets show this simple method can achieve performance similar to that of Gradient Boosted Decision Trees. 
\end{abstract}

\vspace{-.1in}
\section{Introduction}
\vspace{-.1in}

Boosting \cite{adaBoost} and Gradient Boosted Decision Trees (GBDT) \cite{gbdt} are very popular supervised learning methods used in industry. Besides high accuracy, they are fast for making predictions, interpretable and have small memory foot print. However, GBDT training for large datasets is challenging even with highly optimized packages such as XGBoost \cite{xgboost}, LightGBM \cite{light_gbm} or CatBoost \cite{cat_boost}. It is also not possible to incrementally update GBDT models with new data. In contrast, neural networks are compatible with incremental training, GPU acceleration, and end-to-end fine tuning.

There have been several efforts to address tree learning challenges. For example using GPU to speed up the training, \cite{tensorflow_estimator, xgboost_gpu, cat_boost_gpu, light_gbm_gpu}, using structured predictions \cite{Norouzi} or reinforcement learning \cite{reinforce} to find better tree splittings. These, however, further complicate the training process. Another option is to use soft binning or differentiable splitting inside a neural network and simulate a decision tree \cite{dndt, Kontschieder_iccv, Kontschieder_cvpr, Balestriero}. However, these methods are often not scalable for large number of features and deep trees. Finally we can use model distilling, \cite{distil_k, model_compression}, but it requires a trained  model to begin with.

In this paper we first illustrate how to convert a learned decision tree to a single neural network with one hidden layer and an input transformation, similar to \cite{nn_rf_short, nn_rf_long}. We then relax properties of this network such as thresholds and activation functions to train an approximately equivalent decision tree ensemble. The final model, called Hammock, is surprisingly simple. It is a fully connected two layers neural network where the input is quantized and one-hot encoded. Experiments on large and small datasets show this simple method can achieve competitive performance with GBDT models.

\section{Decision Trees and Neural Networks}
\begin{figure}[t]
  \centering
  \includegraphics[width = 400pt]{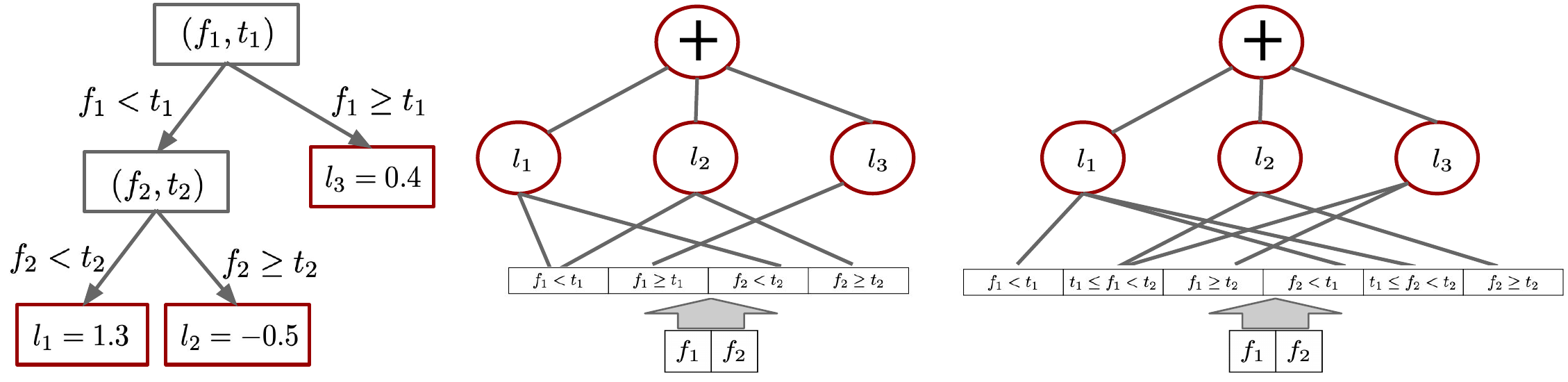}
  \caption{left: a decision tree, middle: decision tree equivalent network, right: Hammock  }
  \label{fig:network_structures}
  \vspace{-.2in}
\end{figure}

\vspace{-.1in}
As shown in figure \ref{fig:network_structures}-left, a decision tree is a collection of rules. For example the tree in this figure is 
\begin{align}
 \mbox{if} & \enskip f_1 < t_1 \enskip \mbox{and} \enskip f_2 < t_2 \quad \Rightarrow 1.3 \nonumber \\
 \mbox{if} & \enskip f_1 < t_1 \enskip \mbox{and} \enskip f_2 \geq t_2 \quad \Rightarrow -0.5 \nonumber \\
 \mbox{if} & \enskip f_1 \geq t_1 \quad \Rightarrow 0.4.
\end{align}

We can implement this rule set using a neural network. One such implementation is shown in figure \ref{fig:network_structures}-middle and consists of an input transformation, a hidden layer where each node represents a tree leaf and an outer node that accumulates outputs of all hidden nodes. The transformation applies all thresholds in the tree nodes to the corresponding features, e.g. transforms feature vector $[f_1, f_2]$ to a binary vector of $[f_1 < t_1, f_1 \geq t1, f_2 < t_2, f_2 \geq t2]$. Weights of the first layer are also binary and are active for a (threshold, leaf) if that pair is in the path between root and the leaf. The bias of the node associated to $i^{th}$ leaf is the number of non-zero weights for that node minus a small number, e.g. $0.1$, and the activation is a step function. The weights between $i^{th}$ hidden node and the outer node is the value of $i^{th}$ leaf. The outer layer can have one node for binary classification or several nodes for the multiclass case. In case of tree ensembles, e.g. GBDT or Random Forest, we can define the input transformation based on all thresholds in all trees and use as many hidden nodes as the total number of leafs in the ensemble. Therefore the whole ensemble can be represented in a single network.

The above network can implement a GBDT but it still requires a trained model. Instead, we can modify it to learn equivalents of tree leafs internally. We start by relaxing all criteria on the network's weights and biases mentioned above. We can then use a pre-determined set of thresholds for each feature and consider all of them in the transformation. For this we can look at each feature distribution and create a set of thresholds, e.g. quantiles.  This is a common practice for speeding up GBDT training process \cite{xgboost}. We can further simplify this transformation by using these thresholds to quantize the feature values and represent the input as one-hot encoding of the quantization bins. This network, called Hammock, is shown in \ref{fig:network_structures}-right. Conceptually, Hammock is a simple network with two fully connected two layers where the input is quantized and one-hot encoded. This can be implemented in TensorFlow using {\textit{feature\_column.bucketized\_column}} \cite{tensorflow}. We can also impose regularization on weights to achieve sparse weights similar to the decision trees. Finally note that because of one-hot encoding after the qunatization, Hammock treats all of its inputs and thresholds as categorical.

\begin{table}[b]
  \begin{center}
    \vspace{-.1in}
    \caption{Accuracy of different methods along with dataset stats}
    \label{tab:results}
\begin{tabular}{lccccccc}
\toprule
{} & XGBoost & Hammock &  LR-NN & NN-1L & \#train &   \#test &  \#feature \\
\midrule
Statlog   & \textbf{0.899} &    0.875 &  0.492 &     0.703 &     4435 &     2000 &         36 \\
Letter    & 0.939 &    \textbf{0.946} &  0.727 &     0.933 &   16000 &     4000 &         16 \\
Optical Digits & 0.955 &    0.947 &  0.943 &     \textbf{0.976} &     3823 &     1797 &         64 \\
Poker      & 0.642 &    \textbf{0.943} &  0.501 &     0.568 &   25010 &  1M &         10 \\
Shuttle   & \textbf{1.000} &    0.999 &  0.925 &     0.998 &    43500 &    14500 &          9 \\
Pen Digits & 0.960 &    0.935 &  0.826 &     \textbf{0.979} &    7494 &     3498 &         16 \\
Netflix Internal & 0.918 & \textbf{0.920} & 0.824 & 0.821 & 10M & 1.5M & 170 \\
\bottomrule
\end{tabular}
  \end{center}
      \vspace{-.1in}
\end{table}

\section{Experiment}
\vspace{-.1in}
We compared performance of Hammock on six UCI datasets for multiclass classification and a large internal dataset for binary classification. In UCI datasets, for Hammock we used $1000$ hidden nodes and $50$ bins for threshold selection, along with $50\%$ dropout and AdaDelta optimizer \cite{ada_delta}. For comparison we used XGBoost (with $100$ trees of max depth $5$), fully connected neural networks with zero and one hidden layer of $1000$ nodes. These are named LR-NN, NN-1L respectively where LR-NN is equivalent to Logistic-Regression and NN-1L is similar to Hammock without the quantization transformation. Table \ref{tab:results} shows accuracy of the models on test data and statistics about each dataset. As shown in this table, Hammock was able to produce competitive results with XGBoost. Compared with NN-L1, Hammock performed better in $5$ out of $7$ cases which shows the advantage of the quantization layer. On Poker dataset, Hammock significantly outperformed XGBoost. A possible explanation is that the decision trees in XGBoost are not not deep enough to capture complexity of the data while Hammock can potentially create a much bigger combination of features in each of its hidden units.

\pagebreak
\bibliographystyle{ieee_fullname}
\bibliography{ref}

\end{document}